\documentclass{article}
\usepackage{iclr2025_re-align_workshop,times}

\usepackage{amsmath,amsfonts,bm}

\def\eqref#1{equation~\ref{#1}}
\def\1{\bm{1}}

\DeclareMathAlphabet{\mathsfit}{\encodingdefault}{\sfdefault}{m}{sl}
\SetMathAlphabet{\mathsfit}{bold}{\encodingdefault}{\sfdefault}{bx}{n}

\usepackage{hyperref}
\usepackage[numbers]{natbib}
\usepackage{url}
\usepackage{graphicx}

\title{Model Connectomes: A Generational \\Approach to Data-Efficient Language Models}

\author{Klemen Kotar \\
Computer Science, Stanford\\
\texttt{klemenk@stanford.edu} \\
\And
Greta Tuckute \\
Brain and Cognitive Sciences, MIT \\
\texttt{gretatu@mit.edu} \\
}

\iclrfinalcopy
\begin{document}

\maketitle

\footnotetext[1]{Code available at: \url{https://github.com/TuKoResearch/GenerationalConnectomes}}

\begin{abstract}
Biological neural networks are shaped both by evolution across generations and by individual learning within an organism’s lifetime, whereas standard artificial neural networks undergo a single, large training procedure without inherited constraints. In this preliminary work, we propose a framework that incorporates this crucial generational dimension—an ``outer loop’’ of evolution that shapes the ``inner loop’’ of learning—so that artificial networks better mirror the effects of evolution and individual learning in biological organisms. Focusing on language, we train a model that inherits a ``model connectome’’ from the outer evolution loop before exposing it to a developmental-scale corpus of 100M tokens. Compared with two closely matched control models, we show that the connectome model performs better or on par on natural language processing tasks as well as alignment to human behavior and brain data. These findings suggest that a model connectome serves as an efficient prior for learning in low-data regimes – narrowing the gap between single-generation artificial models and biologically evolved neural networks.

\end{abstract}
\section{Introduction}
\label{section:intro}
How does the brain quickly and robustly learn to perform a wide array of tasks? A lot of research compares the representations from artificial and biological neural networks to broadly answer this question \cite{sucholutsky2024gettingalignedrepresentationalalignment,tuckute2024language,zador2023catalyzing}. However, artificial and biological networks differ fundamentally across several dimensions, including architecture, input-output constraints, and–critically–their learning processes (e.g., \cite{richards2019deep}). A key distinction lies in how learning takes place over time. In biological systems, learning consists of two nested ``training loops'': The evolutionary outer loop takes place across generations, where information needs to be transmitted in a low-bandwidth, compressed form--such as wiring rules and circuit priors \cite{hinton1987how}. The inner learning loop takes place during the lifetime of a single individual, where relatively little data is used to flexibly learn complex behaviors and generalize to novel settings \cite{zador2019critique,hasson2020direct}. In contrast, standard artificial neural networks rely on a single large-scale training phase with a randomly initialized model using a generic architecture (Transformers; \cite{vaswani_attention_2017}), requiring vast amounts of data \citep{warstadt2022artificial,warstadt2023call}. In this early stage work, we take a step toward bridging this gap between artificial and biological systems by proposing a generational learning framework for neural networks, followed by evaluation of task performance and alignment to human behavior and brain responses.

Our paper focuses on the domain of language, motivated by the massive engineering success of language-trained models \cite{radford2018improving, meta2023llamav3, deepseek2023, kaplan2020scaling}, as well as research that demonstrates alignment of language model representations with human behavior and neural responses during language processing \cite{jain2018incorporating,toneva2019interpreting,schrimpf_integrative_2020,caucheteux2022brains,goldstein2022shared}. 

Our problem formulation consists of two phases: an evolutionary outer loop, where the model has access to a large dataset of super-human scale, and a lifetime learning phase, where the model has to learn efficiently on a smaller, developmental-scale dataset (Figure \ref{fig:main}A). Crucially, any learning acquired from the large dataset must be compressed into a connectome, a sparse binary mask, that is transmitted across ``model generations''. Specifically, across six generations of the evolutionary large-dataset loop--where 20\% of the weights are pruned (i.e., zeroed out) in each generation--we derive a connectome that ultimately retains only 25\% of the initial model weights. Additionally, the remaining weights are set to a fixed positive or negative constant after each generation, further reducing the information capacity of the connectome. This final connectome is then used to initialize a model, denoted as \textbf{Connectome}, which is trained on the smaller, developmental-scale dataset. To evaluate its performance, we compare it to two control models trained on the small dataset: i) \textbf{RandomConnectome}, a model with a randomly sampled connectome that also retains 25\% of the initial weights, and ii) \textbf{NoConnectome}, an unpruned, fully-connected model with uniform initialization (the ``standard choice'' in machine learning) (Figure \ref{fig:main}A).

Our core contribution is demonstrating that distilling information from a large dataset into a sparse connectome for initialization enables a model to generalize more effectively in low-data regimes. The resulting connectome model performs better or on par with control models in natural language processing (NLP) tasks and aligns more closely with human behavior and brain responses.

Related work in machine learning has explored compressing neural network weights by employing lower-bit representations \cite{nagel2022gptq}, enforcing structural sparsity \cite{sanh2020movement}, and routing computations through uniform sub-modules as in mixtures of experts \cite{shazeer2017outrageously}. Another line of research has shown that sparsely connected sub-networks within a larger network, when trained from their original initialization, can match the performance of a full image-trained network \cite{frankle2019lottery}. Subsequent studies have refined this idea through iterative pruning over multiple generations \cite{paul2023unmasking}, extended it to language models \cite{panda2024lottery, zheng2022lottery}, shown its benefits for transfer learning \cite{t2022sparse}, proposed improved pruning policies \cite{tanaka2020pruning}, and demonstrated that sub-networks can be defined using binary masks alone \cite{zhou2019deconstructing}.
We consolidate these ideas under the notion of a \textbf{model connectome} — a sparse binary wiring diagram composed of excitatory and inhibitory connections (i.e., weights with a constant positive or negative sign) that is refined over successive generations. The connectome defines a model’s initialization. While previous work has focused on achieving loss equivalence between pruned and unpruned models on the \textit{same} training dataset (with the goal of reducing model size), our study investigates a pruned model’s (i.e., initialized with a connectome) ability to learn effectively from a much smaller dataset. Hence, in contrast to prior work, our approach highlights the potential of sparse initialization to support learning in data-limited regimes.

One line of related work within ``NeuroAI'' \cite{zador2023catalyzing} has explored directly optimizing the compression of a model's weights to achieve high innate (zero-shot) performance for vision tasks \cite{barabasi2023complex,shuvaev2024encoding}. In contrast, we do not optimize for compression--instead, our connectome evolves over generations through ``standard'' task optimization. In practice, our resulting models converge faster during training, rather than performing well at initialization. 
Another direction has pruned the embedding space of standard pre-trained models to identify model features that are most important for alignment with human behavior or neural data \cite{manrique2023enhancing,truong2024pruning}. Unlike these approaches, we prune our model based solely on next-word prediction performance, assessing human alignment only after generational pruning. To our knowledge, the behavioral and brain alignment of generationally pruned models (here, denoted as ``Connectome'' models) has not previously been studied.

\section{Methods}
\label{methods}
\subsection{Model development}
\subsubsection{Model architecture}
We investigate how language models in the GPT-2 family \cite{radford2018improving} can transmit information through initialization across model generations. We utilize the standard GPT-2 configurations used in modern AI approaches (see Appendix \ref{SI:architecture}). In our main experiments, we train models with 124M weights (see Section \ref{result:nlp}), and, in an exploratory analysis we scale up to 417M weights (see Section \ref{scaling}). To ensure robustness of our results, all analyses are conducted with four seeds per model instantiation.

\subsubsection{Training datasets}
Our framework leverages two datasets: A small ($\boldsymbol{S}$) dataset, approximating a child’s language exposure up to age 10 \cite{hart1992american,warstadt2022artificial,warstadt2023call} (100M tokens of FineWeb \cite{penedo2024the}, $\sim$75M words), and a much larger ($\boldsymbol{L}$) dataset (4B tokens of FineWeb \cite{penedo2024the}, $\sim$3B words) containing roughly the amount of tokens that are optimal for training a standard 124M-parameter GPT model \cite{radford2018improving} (see \ref{SI:word_count} for details). We explore how the $\boldsymbol{L}$ dataset (orders of magnitude larger than $\boldsymbol{S}$) can transmit information through model initialization (``outer'' evolutionary loop), denoted as the model connectome, across successive generations of large language models (LLMs), which are then ultimately trained on the smaller $\boldsymbol{S}$ dataset (``inner'' learning loop). 

\subsubsection{{\textbf{Generational outer loop:} Wiring up the Connectome}}
We define the model connectome as a special form of initialization: a binary mask of weights (``synapses'') connecting LLM units between layers. The connectome has the following two properties: i) \textbf{sparsity}, where most synapses are zeroed out (effectively reducing the size of the final model), and ii) \textbf{binary initialization}, assigning the remaining synapses a binary initialization value; a constant positive (excitatory) or a constant negative (inhibitory). 

To derive the connectome, we explore the Iterative Pruning approach \cite{frankle2019lottery} on the $\boldsymbol{L}$ dataset. We begin with a fully dense normally initialized model with standard parameters ($\mu=0, \sigma=0.02$) \cite{radford2018improving}, $f_{\theta}^{0}$, and train it on $\boldsymbol{L}$ for 7,000 iterations. After training, we take the final state of this model, $f_{\theta}^{0fin}$, and prune (i.e., zero out) $20\%$ of the weights with lowest absolute magnitude, yielding the next model generation’s initialization, $f_{\theta}^{1}$ \footnote{Note that we performed pruning in a layer-wise fashion, as pilot experiment ablations showed that global pruning led to significantly worse performance.}. In addition to pruning, $f_{\theta}^{1}$ retains only the sign of the unpruned weights in $f_{\theta}^{0fin}$, initializing the weights of the subsequent model generation as $0.0$, $-0.02$ or $+0.02$ \cite{zhou2019deconstructing} (where 0.02 is the standard deviation of the random initialization of the initial model parameters). We then train $f_{\theta}^{1}$ for another $7,000$ iterations, only optimizing the remaining $80\%$ unpruned binary initialized weights. This prune-and-retrain cycle is repeated for six generations, ultimately yielding $f_{\theta}^{5fin}$, a model with $25\%$ of the original weights retained from the original model, $f_{\theta}^{0}$. The final $f_\theta^5$ connectome serves as a highly compressed wiring diagram (Appendix \ref{SI:compression}).

For each training generation, we use a batch size of 512 and a learning rate schedule consisting of 250 linear warm-up steps, 5,000 hold steps at 0.0018 followed by 1,750 steps of linear decay to zero as in \cite{modded_nanogpt_2024}. We use the AdamW optimizer and weight decay of 0.1.

\subsubsection{{\textbf{Developmental inner loop:} Learning language representations}}
We take the final connectome, $f_{\theta}^5$, and use it to initialize a new sparse model, which we train on the smaller dataset $\boldsymbol{S}$ (just 100M tokens, different from the $\boldsymbol{L}$ dataset). This model is trained for 2,000 iterations (250 warm-up steps, 1,750 decay steps) using the same batch size, maximum learning rate, weight decay, and optimizer class as in the generational loop. We denote this model as \textbf{Connectome}. We train two control models, also on $\boldsymbol{S}$, to have a set of minimally differing control models: i) \textbf{RandomConnectome}, a model with a similar initialization mask to \textbf{Connectome}, except that the sparsity and the positive/negative weights are randomly sampled, and ii) \textbf{NoConnectome}, a normally initialized unpruned dense model with ($\mu=0, \sigma=0.02$).

\subsection{Model evaluation}
We evaluate models on NLP tasks (Section \ref{result:nlp}), behavioral alignment (Section \ref{results:beh_neural}), and neural alignment (Section \ref{results:beh_neural}), and the respective sections contain brief methods. For detailed methods, see Appendix \ref{SI:modeleval}.

\begin{figure*}[t]
    \centering
    \includegraphics[width=0.8\textwidth]{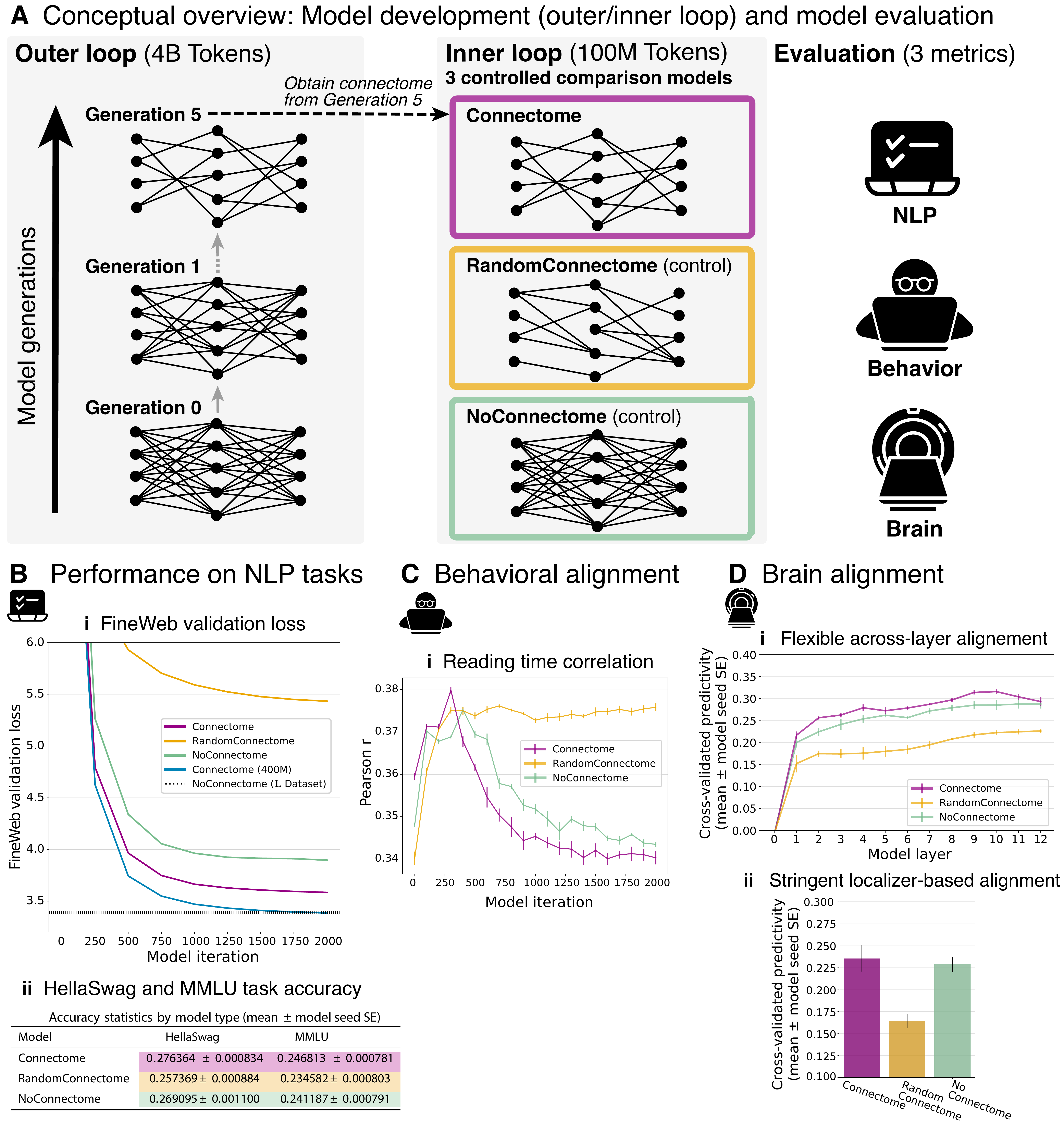}
    \caption{\textbf{A.} Conceptual overview, see description in Sections \ref{section:intro} and \ref{methods}. \textbf{B.} Performance evaluation on standard NLP benchmarks: FineWeb validation loss (panel i), HellaSwag and MMLU (panel ii). \textbf{C.} Alignment with human reading times on naturalistic stories. \textbf{D.} Model-brain alignment, flexibly mapping all units within each model layer to brain responses (panel i) or through a more stringent procedure which localizes language-selective model units (panel ii). To ensure robustness of our results, all analyses are conducted with four seeds per model instantiation, and plots report the standard error of the mean (SE) across seeds.}
    \label{fig:main}
    \vspace{-5mm}
\end{figure*}

\section{Results}
\subsection{The connectome model outperforms control models on NLP benchmarks}
\label{result:nlp}
We evaluate our models on the FineWeb validation loss–-a popular LLM next-word prediction benchmark \cite{penedo2024the}. As shown in Figure \ref{fig:main}B (panel i), the \textbf{Connectome} model strongly outperforms the standard \textbf{NoConnectome} baseline when both are trained on the small dataset $\boldsymbol{S}$, despite \textbf{Connectome} using only $25\%$ of the weights (purple vs. green line). To contextualize the \textbf{Connectome} model's performance, we compare it to an upper bound: a dense model (similarly 124M weights) trained on the full large dataset ($\boldsymbol{L}$) (dotted horizontal line in Figure \ref{fig:main}B). Among the three models trained on the small dataset, \textbf{Connectome} comes closest to this upper bound. Finally, the \textbf{RandomConnectome} model performs substantially worse than both \textbf{Connectome} and \textbf{NoConnectome}, indicating that the iterative pruning procedure is carving out an efficient subspace for learning language. 
Next, we evaluate our models on two standard NLP benchmarks; HellaSwag \cite{zellers2019hellaswag}, a task that requires selecting the most plausible continuation for sentences concerning everyday scenarios, and MMLU \cite{hendrycks2021measuring}, a multiple‐choice benchmark assessing LLMs' knowledge and reasoning on various topics.
In line with the superior performance of \textbf{Connectome} on the FineWeb validation loss (Figure \ref{fig:main}B, panel i), \textbf{Connectome} also outperforms \textbf{RandomConnectome} and \textbf{NoConnectome} on the HellaSwag and MMLU (Figure \ref{fig:main}B, panel ii), highlighting its strong performance as a language model beyond simply next-word prediction. 

\subsubsection{Scaling up: Exploratory analysis}
\label{scaling}
So far (Figure \ref{fig:main}B) we have compared three models (\textbf{Connectome}, \textbf{RandomConnectome}, and \textbf{NoConnectome})--each with 124M weights (with \textbf{Connectome} and \textbf{RandomConnectome} having only $25\%$ active weights, i.e., 31M), and found that the \textbf{Connectome} model consistently outperforms the two control models. But what happens if we take this a step further and scale up? In an exploratory analysis, we initialized a model with 417M weights and pruned it down to 109M active weights over a six-generation procedure similar to that of \textbf{Connectome}. As expected, the 417M pruned model (Figure \ref{fig:main}B, blue line) outperforms the 124M \textbf{Connectome} model (purple line). Interestingly, the 417M pruned model obtains a loss comparable to the 124M dense model trained on the \textit{large} dataset ($\boldsymbol{L}$) (Figure \ref{fig:main}B, dotted line), despite the 417M pruned model being trained only on the \textit{small} dataset ($\boldsymbol{S}$). These findings suggest that a large pruned model--only trained on the small dataset--matches the performance of a small dense model trained on the large dataset, underscoring the data efficiency of the connectome approach.
More broadly, these results demonstrate the power of efficient model initialization through iterative pruning, and open the door for future work scaling to even larger models and evaluating on a larger array of tasks. 
 
\subsection{The Connectome model aligns better or on par with human reading times and brain responses}
\label{results:beh_neural}

\textbf{Behavioral alignment}: Building on prior work demonstrating that per-word surprisal estimates from LLMs correlate with self-paced reading times \cite{wilcox2020predictive,oh2023does,shain2024word} (a measure of language processing difficulty, \cite{hale2001probabilistic,smith2013effect}), we tested our models against a large dataset of reading times from 179 participants reading 5-10 naturalistic stories \cite{futrell2021natural}. We tested the behavioral alignment between the models’ estimated per-word surprisal and the reading time obtained from human participants, as done in prior work \cite{aw2023instruction,alkhamissi2024brain}. Based on Figure \ref{fig:main}C, we note two main findings. First, we replicate the finding that for larger LLMs trained on large amounts of data, alignment with reading times degrades \cite{oh2023transformer,steuer2023large,oh2024frequency,de2023scaling,shain2024word}--here, indicated by a peak behavioral alignment around iteration 250. In line with this finding, the \textbf{RandomConnectome} model, remaining in a high-loss regime, maintains a surprisingly high prediction of reading times. Second, we note that the \textbf{Connectome} model has the highest peak behavioral alignment with human reading times compared to the control models--although this result should be interpreted in light of the inverse correlation of training data amount and fit to human reading times.

\textbf{Brain alignment:} Building on prior work showing that internal LLM representations can predict human brain responses \cite{schrimpf_integrative_2020,caucheteux2022brains,goldstein2022shared,tuckute2024language}, we here evaluated our models’ brain alignment using a published benchmark consisting of responses from the human language network \cite{fedorenko2010new} collected from five participants reading 1,000 linguistically diverse sentences \cite{tuckute2024driving} (see additional details in Appendix \ref{SI:modeleval}). First, we followed the alignment procedure of Tuckute et al. \cite{tuckute2024driving} (Figure \ref{fig:main}D, panel i) fitting a linear encoding model that flexibly maps all units within each LLM layer to brain responses. The \textbf{Connectome} model consistently outperforms the \textbf{RandomConnectome} model while the \textbf{NoConnectome} model is more comparable. The \textbf{Connectome} model achieves \textit{r} = 0.32 for layer 10 (the noise ceiling for this dataset is \textit{r} = 0.56, i.e., 57\%).

Finally, we turn to a more stringent model-brain alignment procedure by focusing on a subset of LLM units that are intended to functionally correspond to the language network in the human brain. To identify these units, we adopt a well-established approach in neuroscience: selecting units that respond more strongly to well-formed sentences than to lists of non-words \cite{fedorenko2010new}. Following AlKhamissi et al. \cite{alkhamissi2024llm} (see details and validation of the approach in this paper), we identify the top 1\% most language-selective units across all layers of each model (yielding 92 units) and apply the same voxelwise encoding procedure used in our previous analyses. This localization step enables a more principled comparison to brain data, with the aim of targeting functionally similar units in models and brains.
Although overall performance drops and there is variability across model seeds, the \textbf{Connectome} model is still more or at least as brain-aligned as the control models (Figure \ref{fig:main}D, panel ii). Similar patterns hold when expanding to the top 10\% language-selective units (Appendix \ref{SI:top10}), with overall higher alignment. 
In conclusion, the \textbf{Connectome} model shows better or on par alignment with brain responses during language processing, highlighting the biological plausibility of our generational modeling framework.

\section{Limitations \& Discussion}
In this paper, we show that a generational learning framework--transmitting sparse connections across model generations--provides an effective model initialization for learning in low-data regimes.
Our \textbf{Connectome} model outperforms control models on NLP tasks and achieves comparable or better alignment with human behavior and brain responses during language processing.

One way to interpret the success of model connectomes is through the lens of model distillation research \cite{hinton2015distilling, chen2020big}, which suggests that large models effectively perform a parallel search over many useful sub-networks during training. The lottery ticket hypothesis \cite{frankle2019lottery, zhou2019deconstructing} builds on this idea, demonstrating that these performant sub-networks can be efficiently extracted from the final model state. We further constrain the problem to a setting where only a binary connectome—capturing the presence and sign of connections—is inherited from an ancestor model trained on a large dataset, and show that this highly compressed structure serves as powerful initializations for efficient learning. This aligns with prior work that has demonstrated the relative importance of sign over magnitude of weight initializations for downstream task performance \cite{zhou2019deconstructing}.
Alternatively, the model connectome can be thought of as a graph, where the outer loop training procedure identifies several good trajectories through which information flows across the model \cite{fernando2017pathnet, han2015learning}. Because these pathways contribute strongly to the model's output, focusing optimization solely on them allows for efficient learning from limited data.

Limitations of this work exist. We by no means claim that our framework mimics the exact process of evolution: Biological organisms do not begin with a dense initialization that ``prunes away'' less important connections over generations (but, excess connections are pruned during development \cite{huttenlocher1979synaptic}). Nevertheless, our framework demonstrates that a highly compressed connectome can transmit sufficient information across generations to evolve a strong initialization, without explicitly optimizing for compression. Another limitation is a limited set of alignment metrics investigated here; future work will explore additional behavioral and brain benchmarks for a more comprehensive view.

Future work includes investigating model behavior across checkpoints in both the outer and inner training loops, scaling up the models (for which we have demonstrated promising results, see Section \ref{scaling}), deploying evolutionary algorithms for pruning (e.g., \cite{wierstra2014natural,stanley2019designing}), and establishing the relationship between the amount of data transmitted through the connectome and performance on various tasks. Additionally, machine learning interpretability techniques (e.g., \cite{wang2022interpretability,marks2024sparse}) could be applied on the connectome to explain specific pruned circuits, and potentially relating them to hypothesized neural circuits.

\subsubsection*{Acknowledgments}
We thank Badr AlKhamissi for sharing the brain-score version of the Futrell2018 behavioral benchmark. We thank Dan Yamins for inspiring conversations. We also thank the Stanford HAI, Stanford Data Sciences and the Marlowe team for computing support. Greta Tuckute acknowledges funding support from MIT's McGovern Institute for Brain Research.

\bibliography{iclr2025_conference}
\bibliographystyle{unsrt}

\appendix
\section{Appendix}

\subsection{Model architecture}
\label{SI:architecture}
We utilize the upgraded GPT-2 architecture which has become the standard baseline implementation for this model family \cite{modded_nanogpt_2024}. It differs from the original GPT-2 paper \cite{radford2018improving} in three ways: firstly it uses RMSNorm without trained parameters \cite{zhang2019rmsnorm}, secondly it removes all bias parameters such that the model consists solely of 2D weight matrices, and thirdly it uses RoPE positional embedding \cite{su2021roformer} instead of learned positional embeddings. These architectural design choices were fixed before evaluating our generational pruning approach. 

\subsection{Dataset details}
\label{SI:word_count}
We utilize the first shard of the FineWeb \cite{penedo2024the} dataset - consisting of 100M tokens which make up 74,248,643 words - for the small dataset $\boldsymbol{S}$, and the subsequent 40 shards - consisting of 4B tokens which make up 2,969,847,300 words - for the large dataset  $\boldsymbol{L}$.

\subsection{Connectome compression}
\label{SI:compression}
Our generational pruning framework iteratively zeros out 20\% of the weights with the lowest magnitudes at each generation. After six generations (see Section \ref{methods}, the final $f_\theta^5$ connectome is a highly compressed representation consisting of just 124M ternary values where $25\%$ are non-zero. The Shannon entropy of this connectome can be computed as: 
$$H = -\left[0.75\log_2(0.75) + 0.125\log_2(0.125) + 0.125\log_2(0.125)\right] = 1.06$$ 
which means that encoding 1.06 bits of information per weight would result in a final compressed size of roughly $124M * 1.6$ bits = 131.44 Mb = 16MB. This corresponds to approximately a 15x compression over naïvely storing the dense model weights (248MB at 16 bits per weight), and an over 500x compression of the entire dataset $\boldsymbol{L}$ (8GB at 16 bits per token).

\subsection{Model evaluation detailed methods}
\label{SI:modeleval}
\subsubsection{Natural language processing tasks}
We evaluate our models on three standard NLP benchmarks: FineWeb validation loss \cite{penedo2024the}, a diverse corpus of web-based text spanning various online content, HellaSwag \cite{zellers2019hellaswag}, a benchmark that tests for commonsense reasoning in scenario completion, and MMLU \cite{hendrycks2021measuring}, a multiple choice benchmark which tests for multi-domain knowledge and reasoning.

\subsubsection{Behavioral alignment}
We analyze the self-paced reading dataset from Futrell et al. (2021) \cite{futrell2021natural} (through Brain-Score \cite{schrimpf2018brain}) consisting of word-by-word reading times from 179 participants across 5-10 naturalistic stories. The same stories are processed by our language models, and behavioral alignment is measured using the Pearson correlation between LLM per-word perplexity (summing sub-token perplexities for words that are split into multiple tokens per \cite{doi:10.1073/pnas.2105646118}) and human reading times.

\subsubsection{Brain alignment}
We analyze fMRI data from Tuckute et al. (2024) \cite{tuckute2024driving} consisting of brain responses from 5 participants during a sentence-reading experiment. Participants read 1,000 6-word-long semantically and stylistically diverse sentences, and we investigated responses in the left-hemisphere language network \cite{fedorenko2010new}, averaged across 5 participants. Following \cite{tuckute2024driving}, brain alignment is measured by predicting brain responses using a ridge regression encoding model and computing the Pearson correlation between predicted and actual responses via 5-fold cross-validation.

\subsection{Brain alignment using top 10\% language-selective units}
\label{SI:top10}

\begin{figure*}[h]
    \centering
    \includegraphics[width=0.45\textwidth]{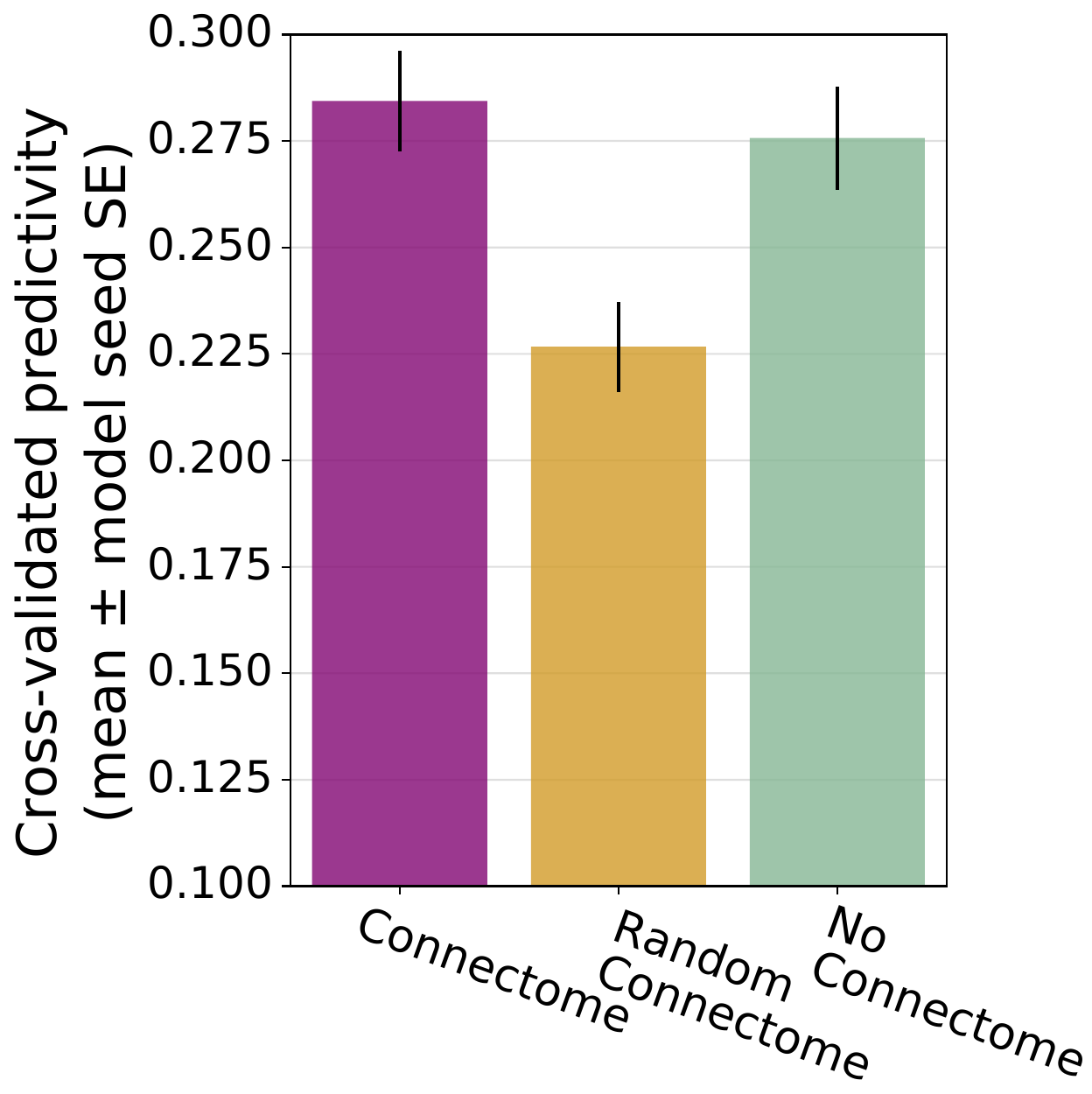}
    \caption{In the main text, we present model-brain alignment results using the top 1\% language-selective units, per prior work \cite{alkhamissi2024llm} (Figure \ref{fig:main}B, panel ii). However, in neuroscience, the top 10\% units are typically used \cite{fedorenko2010new}, and in this supplement we select the top 10\% language-selective units in our models. The pattern is the same as in the top 1\% case (Figure \ref{fig:main}D, panel ii), but the overall correlations are higher. The error bars show show the standard error of the mean (SE) across model seeds.}
\end{figure*}

\end{document}